\documentclass{article}

\usepackage[final]{nips_2017}

\usepackage[utf8]{inputenc} 
\usepackage[T1]{fontenc}    
\usepackage{hyperref}       
\usepackage{url}            
\usepackage{booktabs}       
\usepackage{amsmath}
\usepackage{amsfonts}       
\usepackage{nicefrac}       
\usepackage{microtype}      
\usepackage{natbib}
\usepackage{graphicx}
\usepackage{paralist}
\usepackage{tabularx}
\usepackage{color}

\usepackage{blindtext}
\usepackage{scrextend}
\addtokomafont{labelinglabel}{\sffamily}

\usepackage[subtle]{savetrees}

\DeclareMathOperator*{\argmax}{arg\,max}
\newcommand{\defeq}{\stackrel{\text{def}}{=}}

\title{A Deep Reinforcement Learning Chatbot \\ (Short Version)}

\author{
  \normalfont \textbf{Iulian V. Serban}, \textbf{Chinnadhurai Sankar}, \textbf{Mathieu Germain}, \textbf{Saizheng Zhang}, \textbf{Zhouhan Lin}, \\ \textbf{Sandeep Subramanian},
  \textbf{Taesup Kim}, \textbf{Michael Pieper}, \textbf{Sarath Chandar}, \textbf{Nan Rosemary Ke}, \\ \textbf{Sai Rajeswar}, \textbf{Alexandre de Brebisson},
  \textbf{Jose M. R. Sotelo}, \textbf{Dendi Suhubdy}, \\ \textbf{Vincent Michalski}, \textbf{Alexandre Nguyen}, \textbf{Joelle Pineau}$^{1,2}$ and \textbf{Yoshua Bengio}$^2$ \\
  Montreal Institute for Learning Algorithms, Montreal, Quebec, Canada
}

\begin{document}

\maketitle

\begin{abstract} \mbox{} \\
We present MILABOT: a deep reinforcement learning chatbot developed by the Montreal Institute for Learning Algorithms (MILA) for the Amazon Alexa Prize competition.
MILABOT is capable of conversing with humans on popular small talk topics through both speech and text.
The system consists of an ensemble of natural language generation and retrieval models, including neural network and template-based models.
By applying reinforcement learning to crowdsourced data and real-world user interactions, the system has been trained to select an appropriate response from the models in its ensemble.
The system has been evaluated through A/B testing with real-world users, where it performed significantly better than other systems.
The results highlight the potential of coupling ensemble systems with deep reinforcement learning as a fruitful path for developing real-world, open-domain conversational agents.
\end{abstract}

\stepcounter{footnote}
\footnotetext{School of Computer Science, McGill University.}
\stepcounter{footnote}
\footnotetext{CIFAR Fellow.}

\section{Introduction}
Conversational agents - including chatbots and personal assistants - are becoming increasingly ubiquitous.
In 2016, Amazon proposed an international university competition with the goal of building a socialbot: a spoken conversational agent capable of conversing with humans on popular topics, such as entertainment, fashion, politics, sports, and technology.\footnote{See \url{https://developer.amazon.com/alexaprize}.}
This article describes the experiments by the \textit{MILA Team} at University of Montreal, with an emphasis on reinforcement learning.

Our socialbot is based on a large-scale ensemble system leveraging deep learning and reinforcement learning.
The ensemble consists of deep learning models, template-based models and external API webservices for natural language retrieval and generation. 
We apply reinforcement learning --- including value function and policy gradient methods --- to intelligently combine an ensemble of retrieval and generation models.
In particular, we propose a novel off-policy model-based reinforcement learning procedure, which yields substantial improvements in A/B testing experiments with real-world users. 

On a rating scale $1-5$, our best performing system reached an average user score of $3.15$, while the average user score for all teams in the competition was only $2.92$.\footnote{Throughout the semi-finals, we carried out several A/B testing experiments to evaluate different variants of our system (see Section \ref{sec:ab_testing_experiments}). The score $3.15$ is based on the best performing system in these experiments.} 
Furthermore, our best performing system averaged $14.5-16.0$ turns per conversation, which is significantly higher than all other systems. 

As shown in the A/B testing experiments, a key ingredient to achieving this performance is the application of off-policy deep reinforcement learning coupled with inductive biases, designed to improve the system's generalization ability by making a more efficient bias-variance tradeoff.


\section{System Overview}\label{sec:dialogue_system_overview}

Early work on dialogue systems~\citep{weizenbaum1966eliza,aust1995philips,mcglashan1992dialogue,simpson1993black} were based mainly on states and rules hand-crafted by human experts. 
Modern dialogue systems typically follow a hybrid architecture, which combines hand-crafted states and rules with statistical machine learning algorithms ~\citep{suendermann2015halef}. 
Due to the complexity of human language, however, it is impossible to enumerate all of the states and rules required for building a socialbot capable of conversing with humans on open-domain, popular topics.
In contrast to such rule-based systems, our core approach is built entirely on statistical machine learning.
We believe that this is the most plausible path to artificially intelligent conversational agents.
The system architecture we propose aims to make as few assumptions as possible about the process of understanding and generating natural language.
As such, the system utilizes only a small number of hand-crafted states and rules.
Meanwhile, every system component has been designed to be optimized (trained) using machine learning algorithms. 
By optimizing these system components first independently on massive datasets and then jointly on real-world user interactions, the system will learn implicitly all relevant states and rules for conducting open-domain conversations.
Given an adequate amount of examples, such a system should outperform any system based on states and rules hand-crafted by human experts.
Further, the system will continue to improve in perpetuity with additional data.

\begin{figure}[h]
  \centering
  \includegraphics[scale=0.25]{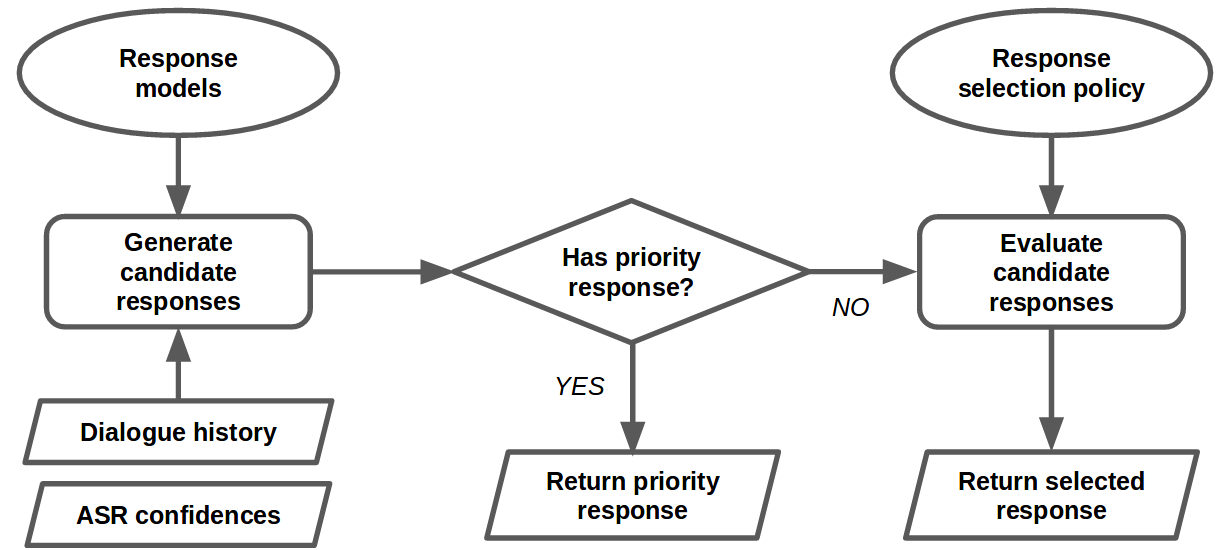}
  \caption{Dialogue manager control flow.}
  \label{fig:dialogue_system}
\end{figure}

Our system architecture is inspired by the success of ensemble-based machine learning systems.
These systems consist of many independent sub-models combined intelligently together.
Examples of such ensemble systems include the winner of the Netflix Prize~\citep{koren2009matrix}, the IBM Watson question-answering system~\citep{ferrucci2010building} and Google's machine translation system~\citep{wu2016google}.

Our system consists of an ensemble of response models (see Figure \ref{fig:dialogue_system}).
Each response model takes as input a dialogue history and outputs a response in natural language text.
As will be explained later, the response models have been engineered to generate responses on a diverse set of topics using a variety of strategies.
The \textit{dialogue manager} is responsible for combining the response models together.
As input, the dialogue manager expects to be given a dialogue history (i.e.\@ all utterances recorded in the dialogue so far, including the current user utterance) and confidence values of the automatic speech recognition system (ASR confidences).
To generate a response, the dialogue manager follows a three-step procedure.
First, it uses all response models to generate a set of candidate responses.
Second, if there exists a \textit{priority} response in the set of candidate responses (i.e.\@ a response which takes precedence over other responses), this response will be returned by the system. 
For example, for the question \textit{"What is your name?"}, the response \textit{"I am an Alexa Prize socialbot"} is a priority response.
Third, if there are no priority responses, the response is selected by the \textit{model selection policy}.
For example, the model selection policy may select a response by scoring all candidate responses and picking the highest-scored response.
\newcolumntype{L}[1]{>{\raggedright\arraybackslash}p{#1}}

\begin{table*}[ht]
 \caption{Example dialogues and candidate responses generated by response models. The chosen system response is marked in bold.} \medskip
 \label{table:candidate_response-examples}
 \small
 \centering 
 \begin{tabular}{L{58mm}L{72mm}} \toprule
 \textbf{Dialogue} & \textbf{Dialogue} \\ \hline
    SYSTEM: Hi. This is an Alexa Prize socialbot. \newline USER: Hi \newline SYSTEM: Do you have pets? \newline USER: Yes, I have two rabbits. & ... \newline USER: Let's talk about movies. \newline SYSTEM: What's your favorite movie? \newline USER: star wars movie \\ \\
     \textbf{Candidate Responses} & \textbf{Candidate Responses} \\ \hline
    BoWFactGenerator: Did you know that male rabbits are called bucks, females are does? \newline GRUQuestionGenerator: How about the American family of it? \newline Initiatorbot: Do you believe in love at first sight? \newline LSTMClassifierMSMarco: Mar 23, 2017  Two Parts: Learning What to Feed Your Rabbit Learning How to ... \newline DualEncoderRedditNews: bye doggie \newline \textbf{Alicebot: Hurrah! Two is a good number of rabbits.} \newline Elizabot: OK, but can you elaborate a bit? \newline  VHREDRedditMovies: Hmm. Nope, never heard of it. \newline VHREDSubtitles: Would you like to see them? \newline  BoWWashingtonPost: I do but I call people with dogs crazy also. &  BoWFactGenerator: Here's an interesting fact. The international telephone dialing code for Antarctica is 672. \newline GRUQuestionGenerator: how about your New York City? \newline Initiatorbot: How was your day? \newline LSTMClassifierMSMarco: A third Anthology film will be released in 2020. \newline DualEncoderRedditNews: Now whisper it to me , one more time bby. \newline Alicebot: What did you like about the robots in that movie? \newline Elizabot: How does that make you feel? \newline \textbf{Evi: Star Wars movie a movie in the Star Wars series.}   \newline VHREDRedditMovies: Oh please. Please. Pleeeease. Let this happen. \newline VHREDSubtitles: What? \newline BoWWashingtonPost: A much more enjoyable feature than last year's old-timer's convention masquerading as a star wars movie. \\
 \bottomrule
 \end{tabular}
\end{table*}

\section{Response Models}
There are 22 response models in the system, including neural network based retrieval models, neural network based generative models, knowledge base question answering systems and template-based systems.
Examples of candidate model responses are shown in Table \ref{table:candidate_response-examples} along with the model names.
For a description of these models, the reader is referred to the technical report by \citet{serban2017deep}.

\section{Model Selection Policy}
After generating the candidate response set, the dialogue manager uses a \textit{model selection policy} to select the response it returns to the user.
The dialogue manager must select a response which increases the satisfaction of the user for the entire dialogue.
In order to do this, it must make a trade-off between immediate and long-term user satisfaction.
For example, suppose the user asks to talk about politics.
If the dialogue manager chooses to respond with a political joke, the user may be pleased for one turn.
Afterwards, however, the user may be disappointed with the system's inability to debate political topics.
Instead, if the dialogue manager chooses to respond with a short news statement, the user may be less pleased for one turn.
However, this may influence the user to follow up with factual questions, which the system may be better adept at handling.
To make the trade-off between immediate and long-term user satisfaction, we consider selecting the appropriate response as a \textit{sequential decision making problem}.
This section describes the five approaches we have investigated to learn the model selection policy. The approaches are evaluated with real-world users in the next section.

We use the reinforcement learning framework~\citep{sutton1998reinforcement}.
The dialogue manager is an agent, which takes actions in an environment in order to maximize rewards.
For each time step $t=1,\dots,T$, the agent observes the dialogue history $h_t$ and must choose one of $K$ actions (responses): $a_t^1,\dots,a_t^K$.
After taking an action, it receives a reward $r_t$ and is transferred to the next state $h_{t+1}$ (which includes the action and the user's next response) where it is provided with a new set of $K$ actions: $a_{t+1}^1,\dots,a_{t+1}^K$.
The agent must maximize the discounted sum of rewards, $R = \sum_{t=1}^T \gamma^t r_t$, where $\gamma \in (0,1]$ is a discount factor.

\textbf{Action-value Parametrization}: We use two different approaches to parametrize the agent's policy.
The first approach is based on an action-value function, defined by parameters $\theta$:
\begin{align}
Q_{\theta}(h_t, a_t^k) \in \mathbb{R} \quad \text{for} \ k=1,\dots,K, \label{eq:action_value_function}
\end{align}
which estimates the expected discounted sum of rewards -- referred to as the \textit{expected return} -- of taking action $a_t^k$ (candidate response $k$) given dialogue history $h_t$ and given that the agent will continue to use the same policy afterwards.
Given $Q_{\theta}$, the agent chooses the action with highest expected return:
\begin{align}
\pi_{\theta}(h_t) = \argmax_{k} \ Q_{\theta}(h_t, a_t^k). \label{eq:greedy_action_value_function}
\end{align}
This approach is related to recent work by \citet{lowe2017toward} and \citet{yu2016strategy}.

\textbf{Stochastic Policy Parametrization}: This approach instead parameterizes a distribution over actions:
\begin{align}
\pi_{\theta}(a_t^k | h_t) = \dfrac{ e^{\lambda^{-1}f_{\theta}(h_t, a_t^k)}}{ \sum_{a_t'}  e^{\lambda^{-1}f_{\theta}(h_t, a_t')}} \quad \text{for} \ k=1,\dots,K, \label{eq:stochastic_policy}
\end{align}
where $f_{\theta}(h_t, a_t^k)$ is the \textit{scoring function}, parametrized by $\theta$, which assigns a scalar score to each response $a_t^k$ conditioned on $h_t$.
The parameter $\lambda$ controls the entropy of the distribution.
The stochastic policy can be transformed to a deterministic (greedy) policy by selecting the action with highest probability:
\begin{align}
\pi_{\theta}^{\text{greedy}}(h_t) = \argmax_{k} \ \pi_{\theta}(a_t^k | h_t). \label{eq:greedy_stochastic_policy}
\end{align}

We parametrize the scoring function and action-value function as neural networks with five layers. 
The first layer is the input, consisting of 1458 features representing both the dialogue history, $h_t$, and the candidate response, $a_t$.
These features are based on a combination of word embeddings, dialogue acts, part-of-speech tags, unigram word overlap, bigram word overlap and model-specific features.\footnote{To limit the effect of speech recognition errors in our experiments, ASR confidence features are not included.} 
The second layer contains 500 hidden units, computed by applying a linear transformation followed by the rectified linear activation function to the input layer features. 
The third layer contains 20 hidden units, computed by applying a linear transformation to the preceding layer units.
The fourth layer contains 5 outputs units, which are probabilities (i.e.\@ all values are positive and their sum equals one).
These output units are computed by applying a linear transformation to the preceding layer units followed by a softmax transformation.
This layer corresponds to the Amazon Mechanical Turk labels, described later.
The fifth layer is the final output layer, which is a single scalar value computed by applying a linear transformation to the units in the third and fourth layers.
In order to learn the parameters, we use five different machine learning approaches described next.

\textbf{Supervised Learning with Crowdsourced Labels}: 
The first approach to learning the policy parameters is called \emph{Supervised Learning AMT}.
This approach  estimates the action-value function $Q_\theta$ using supervised learning on crowdsourced labels.
It also serves as initialization for all other approaches.

We use Amazon Mechanical Turk (AMT) to collect data for training the policy.
We follow a setup similar to \citet{liu2016not}.
We show human evaluators a dialogue along with 4 candidate responses, and ask them to score how appropriate each candidate response is on a 1-5 Likert-type scale.
The score 1 indicates that the response is inappropriate or does not make sense, 3 indicates that the response is acceptable, and 5 indicates that the response is excellent and highly appropriate.
As examples, we use a few thousand dialogues recorded between Alexa users and a preliminary version of the systems.
The corresponding candidate responses are generated by the response models.
In total, we collected 199,678 labels,
which are split this into training (train), development (dev) and testing (test) datasets consisting of respectively 137,549, 23,298 and 38,831 labels each.

We optimize the model parameters $\theta$ w.r.t.\@ log-likelihood (cross-entropy) using mini-batch stochastic gradient descent (SGD) to predict the 4th layer, which represents the AMT labels.
Since we do not have labels for the last layer of the model, we fix the corresponding linear transformation parameters to $[1.0, 2.0, 3.0, 4.0, 5.0]$.
In this case, we assign a score of $1.0$ for an inappropriate response, $3.0$ for an acceptable response and $5.0$ for an excellent response.

\newpage
\textbf{Off-policy REINFORCE}: 
Our next approach learns a stochastic policy directly from examples of dialogues recorded between the system and real-world users.
Let $\{h_t^d, a_t^d, R^d\}_{d, t}$ be a set of examples, where $d$ indicates the dialogue and $t$ indicates the time step (turn).
Let $h_t^d$ be the dialogue history, $a_t^d$ be the agent's action and $R^d$ be the observed return.
Further, let $\theta_d$ be the parameters of the stochastic policy $\pi_{\theta_t}$ used during dialogue $d$.
We use a re-weighted variant of the \textit{REINFORCE} algorithm~\citep{williams1992simple,precup2000eligibility,precup2001off}, with learning rate $\alpha > 0$, which updates the policy parameters for each example $(h_t^d, a_t^d, R^d)$:
\begin{align}
\Delta \theta \ = \ \alpha \dfrac{\pi_{\theta}(a_{t}^d | h_{t}^d)}{\pi_{\theta_d}(a_{t}^d | h_{t}^d)} \ \nabla_{\theta} \ \log \  \pi_{\theta}(a_t^d | h_t^d) \ R^d. \label{eq:offpolicy_reinforce}
\end{align}
The intuition behind the algorithm is analogous to learning from trial and error.
Examples with high user scores $R^d$ will increase the probability of the actions taken by the agent through the term $\nabla_{\theta} \ \log \  \pi_{\theta}(a_t^d | h_t^d) \ R^d$.
Vice versa, examples with low user scores will decrease the action probabilities. 
The ratio on the left-hand-side corrects for the discrepancy between the learned policy, $\pi_{\theta}$, and the policy under which the data was collected, $\pi_{\theta_d}$.
We evaluate the policy using an estimate of the expected return:
\begin{align}
 \text{E}_{\pi_{\theta}}[R] \ \approx \ \sum_{d,t} \dfrac{\pi_{\theta}(a_{t}^d | h_{t}^d)}{\pi_{\theta_d}(a_{t}^d | h_{t}^d)} \ R^d. \label{eq:offpolicy_reinforce_evaluation}
\end{align}

For training, we use over five thousand dialogues and scores collected in interactions between real users and a preliminary version of our system between June 30th to July 24th, 2017.
We optimize the policy parameters on a training set with SGD based on eq.\@ \eqref{eq:offpolicy_reinforce}.
We select hyper-parameters and early-stop on a development set based on eq.\@ \eqref{eq:offpolicy_reinforce_evaluation}.

\textbf{Learned Reward Function}: 
Our two next approaches trains a linear regression model to predict the user score from a given dialogue.
Given a dialogue history $h_t$ and a candidate response $a_t$, the model $g_{\phi}$, with parameters $\phi$, predicts the corresponding user score.
As training data is scarce, we only use 23 higher-level features as input.
The model is trained on the same dataset as \emph{Off-policy REINFORCE}.

The regression model $g_{\phi}$ is used in two ways.
First, it is used to fine-tune the action-value function learned by \emph{Supervised Learning AMT} to more accurately predict the user score. 
Specifically, the output layer is fine-tuned w.r.t.\@ the squared-error between its own prediction and $g_{\phi}$.
This new policy is called \emph{Supervised AMT Learned Reward}.
Second, the regression model is combined with \textit{Off-policy REINFORCE} into a policy called \textit{Off-policy REINFORCE Learned Reward}.
This policy is trained as \textit{Off-policy REINFORCE}, but where $R^d$ is replaced with the predicted user score $g_{\phi}$ in eq.\@ \eqref{eq:offpolicy_reinforce}.

\textbf{Q-learning with the Abstract Discourse Markov Decision Process}:
Our final approach is based on learning a policy through a simplified Markov decision process (MDP), called the \emph{Abstract Discourse MDP}.
This approach is somewhat similar to training with a user simulator.
The MDP is fitted on the same dataset of dialogues and user scores as before.
In particular, the per time-step reward function of the MDP is set to the score predicted by \emph{Supervised AMT}.
For a description of the MDP, the reader is referred to the technical report by~\citet{serban2017deep}.

Given the \emph{Abstract Discourse MDP}, we use \textit{Q-learning} with \textit{experience replay} to learn the policy with an action-value parametrization~\citep{mnih2013playing,lin1993reinforcement}.
We use an experience replay memory buffer of size $1000$ and an $\epsilon$-greedy exploration scheme with $\epsilon=0.1$.
We experiment with discount factors $\gamma \in \{0.1, 0.2, 0.5\}$.
Training is based on SGD and carried out in two alternating phases.
For every $100$ episodes of training, we evaluate the policy over $100$ episodes w.r.t.\@ average return.
During evaluation, the dialogue histories are sampled from a separate set of dialogue histories.
This ensures that the policy is not \textit{overfitting} the finite set of dialogue histories.
We select the policy which performs best w.r.t.\@ average return.
This policy is called \emph{Q-learning AMT}.
A quantitative analysis shows that the learned policy is more likely to select \textit{risky} responses, perhaps because it has learned effective \textit{remediation} or \textit{fall-back} strategies ~\citep{serban2017deep}.

\section{A/B Testing Experiments} \label{sec:ab_testing_experiments}
We carry out A/B testing experiments to evaluate the dialogue manager policies for selecting the response model.
When an Alexa user starts a conversation with the system, they are assigned at random to a policy and afterwards the dialogue and their score is recorded.

A major issue with the A/B testing experiments is that the distribution of Alexa users changes through time.
Different types of users will be using the system depending on the time of day, weekday and holiday season.
In addition, user expectations towards our system change as users interact with other socialbots in the competition.
Therefore, we must take steps to reduce confounding factors and correlations between users.
First, during each A/B testing experiment, we simultaneously evaluate all policies of interest.
This ensures that we have approximately the same number of users interacting with each policy w.r.t.\@ time of day and weekday.
This minimizes the effect of the changing user distribution \textit{within} each A/B testing period.
Second, we discard scores from returning users (i.e.\@ users who have already evaluated the system once).
Users who are returning to the system are likely influenced by their previous interactions with the system.
For example, users who had a positive previous experience may be biased towards giving higher scores in their next interaction.

\subsection{Experiment Periods}
\textbf{Exp \#1:}
The first A/B testing experiment was conducted between July 29th and August 6th, 2017.
We tested the dialogue manager policies \emph{Supervised AMT}, \emph{Supervised AMT Learned Reward}, \emph{Off-policy REINFORCE}, \emph{Off-policy REINFORCE Learned Reward} and \emph{Q-learning AMT}.
We used the greedy variants for the Off-policy REINFORCE policies. 
We also tested a heuristic baseline policy \emph{Evibot + Alicebot}, which selects the \emph{Evibot} model response if available, and otherwise selects the \emph{Alicebot} model response.
Over a thousand user scores were collected with about two hundred user scores per policy.\footnote{To ensure high accuracy, human annotators were used to transcribe the audio related to the Alexa user scores for the first and second experiments. Amazon's speech recognition system was used in the third experiment.}

This experiment occurred in the middle of the competition semi-finals.
In this period, users are likely to have relatively few expectations towards the systems in the competition (e.g.\@ that the system can converse on a particular topic or engage in \textit{non-conversational activities}, such as playing games).
Further, the period July 29th - August 6th overlaps with the summer holidays in the United States.
As such, we might expect more children to interact with system here than during other seasons.

\textbf{Exp \#2:}
The second A/B testing experiment was conducted between August 6th and August 15th, 2017.
We tested the two policies \emph{Off-policy REINFORCE} and \emph{Q-learning AMT}.
Prior to beginning the experiment, minor system improvements were carried out w.r.t.\@ the \emph{Initiatorbot} and filtering out profanities.
In total, about six hundred user scores were collected per policy.

This experiment occurred at the end of the competition semi-finals.
At this point, many users have already interacted with other socialbots in the competition, and are therefore likely to have developed expectations towards the systems (e.g.\@ conversing on a particular topic or engaging in \textit{non-conversational activities}, such as playing games).
Further, the period August 6th - August 15th overlaps with the end of the summer holidays and the beginning of the school year in the United States.
This means we should expect less children interacting than in the previous A/B testing experiment.

\textbf{Exp \#3:}
The third A/B testing experiment was carried out between August 15th, 2017 and August 21st, 2017.
Due to the surprising results in the previous A/B testing experiment, we decided to continue testing the two policies \emph{Off-policy REINFORCE} and \emph{Q-learning AMT}.
In total, about three hundred user ratings were collected after discarding returning users.

This experiment occurred after the end of the competition semi-finals.
This means that it is likely that many Alexa users have already developed expectations towards the systems.
Further, the period August 15th - August 21st lies entirely within the beginning of the school year in the United States.
We might expect less children to interact with the system than in the previous A/B testing experiment.

\begin{table}[t]
  \caption{A/B testing results ($\pm$ 95\% confidence intervals). Stars indicate statistical significance at $95\%$.} \label{tabel:ab_testing_round_one}
  \small
  \centering
    \begin{tabular}{llcccc}
     \toprule
     & \textbf{Policy} & \textbf{User score} & \textbf{Dialogue length} & \textbf{Pos.\@ utterances} & \textbf{Neg.\@ utterances} \\
    \midrule
    \textbf{Exp \#1} & \emph{Evibot + Alicebot} & $2.86 \pm 0.22 $ & $ 31.84 \pm 6.02 $ & $2.80\% \pm 0.79 $ & $5.63\% \pm 1.27$ \\
    & \emph{Supervised AMT} & $ 2.80 \pm 0.21 $ & $ 34.94 \pm 8.07 $ & $\mathbf{4.00\% \pm 1.05}$ & $8.06\% \pm 1.38$ \\
    & \parbox[c][2.65em][c]{0.225\textwidth}{\emph{Supervised AMT} \\ \protect{\hphantom{\ }} \emph{Learned Reward}} & $2.74 \pm 0.21 $ & $ 27.83 \pm 5.05 $ & $2.56\% \pm 0.70 $ & $6.46\% \pm 1.29$ \\
    & \emph{Off-policy REINFORCE} & $2.86 \pm 0.21 $ & $\mathbf{37.51 \pm 7.21}$ & $3.98\% \pm 0.80 $ & $6.25 \pm 1.28$ \\
    & \parbox[c][2.65em][c]{0.225\textwidth}{\emph{Off-policy REINFORCE} \\ \protect{\hphantom{\ }} \emph{Learned Reward}} & $2.84 \pm 0.23 $ & $ 34.56 \pm 11.55 $ & $ 2.79\% \pm 0.76$ & $ 6.90\% \pm 1.45$ \\
    & \emph{Q-learning AMT}* & $\mathbf{3.15 \pm 0.20}$ & $ 30.26 \pm 4.64 $ & $3.75\% \pm 0.93 $ & $\mathbf{5.41\% \pm 1.16}$ \\ \midrule
    \textbf{Exp \#2} & \emph{Off-policy REINFORCE} & $\mathbf{3.06 \pm 0.12}$ & $\mathbf{34.45 \pm 3.76}$ & $3.23\% \pm 0.45$ & $7.97\% \pm 0.85$ \\
     & \emph{Q-learning AMT} & $2.92 \pm 0.12$ & $31.84 \pm 3.69$ & $\mathbf{3.38\% \pm 0.50}$ & $\mathbf{7.61\% \pm 0.84}$ \\ \midrule
    \textbf{Exp \#3} & \emph{Off-policy REINFORCE} & $3.03 \pm 0.18$ & $30.93 \pm 4.96$ & $2.72 \pm 0.59$ & $7.36 \pm 1.22$ \\
    & \emph{Q-learning AMT} & $\mathbf{3.06 \pm 0.17}$ & $\mathbf{33.69 \pm 5.84}$ & $\mathbf{3.63 \pm 0.68}$ & $\mathbf{6.67 \pm 0.98}$     \\ \bottomrule
    \end{tabular}
\end{table}

\subsection{Results \& Discussion}


Table \ref{tabel:ab_testing_round_one} shows the average Alexa user scores and average dialogue length, as well as average percentage of positive and negative user utterances according to a sentiment classifier.\footnote{$95\%$ confidence intervals are computed under the assumption that the Alexa user scores for each policy are drawn from a normal distribution with its own mean and variance.}

We observe that \emph{Q-learning AMT} performed best among all policies w.r.t.\@ Alexa user scores in the first and third experiments.
In the first experiment, \emph{Q-learning AMT} obtained an average user score of $3.15$, which is significantly better than all other policies at a $95\%$ significance level under a two-sample t-test.
This is supported by the percentage of user utterances with positive and negative sentiment, where \emph{Q-learning AMT} consistently obtained the lowest percentage of negative sentiment user utterances while maintaining a high percentage of positive sentiment user utterances.
In comparison, the average user score for all the teams in the competition during the semi-finals was only $2.92$.
Next comes \emph{Off-policy REINFORCE}, which performed best in the second experiment.
In the second and third experiments, \emph{Off-policy REINFORCE} also performed substantially better than all the other policies in the first experiment.
Further, in the first experiment, \emph{Off-policy REINFORCE} also achieved the longest dialogues with an average of $37.51/2 = 18.76$ turns per dialogue.
In comparison, the average number of turns per dialogue for all the teams in the competition during the semi-finals was only $11$.\footnote{This number was reported by Amazon.}
This means \emph{Off-policy REINFORCE} has over $70\%$ more turns on average than the other teams in the competition semi-finals.
This is remarkable since it does not utilize \textit{non-conversational activities} and has few negative user utterances.
The remaining policies achieved average user scores between $2.74$ and $2.86$, suggesting that they have not learned to select responses more appropriately than the heuristic policy \emph{Evibot + Alicebot}.

In addition, we computed several linguistic statistics for the policies in the first experiment. On average, the \emph{Q-learning AMT} responses contained 1.98 noun phrases, while the \emph{Off-policy REINFORCE} and \emph{Evibot + Alicebot} responses contained only 1.45 and 1.05 noun phrases respectively.
Further, on average, the \emph{Q-learning AMT} responses had a word overlap with their immediate preceding user utterances of 11.28, while the \emph{Off-policy REINFORCE} and \emph{Evibot + Alicebot} responses had a word overlap of only 9.05 and 7.33 respectively.
This suggests that \emph{Q-learning AMT} has substantially more topical specificity (semantic content) and topical coherence (likelihood of staying on topic) compared to all other policies. As such, it seems likely that returning users would prefer this policy over others. This finding is consistent with the analysis showing that \emph{Q-learning AMT} is more \textit{risk tolerant}.

In conclusion, the two policies \emph{Q-learning AMT} and \emph{Off-policy REINFORCE} have demonstrated substantial improvements over all other policies. 
Further, the \emph{Q-learning AMT} policy achieved an average Alexa user score substantially above the average of the all teams in the Amazon Alexa competition semi-finals.
This strongly suggests that learning a policy through simulations in an \emph{Abstract Discourse MDP} may serve as a fruitful path towards developing open-domain socialbots.
The performance of \emph{Off-policy REINFORCE} suggests that optimizing the policy directly towards user scores also may serve as a fruitful path.
In particular, \emph{Off-policy REINFORCE} obtained a substantial increase in the average number of turns in the dialogue compared to the average of all teams in the semi-finals, suggesting that the resulting conversations are significantly more interactive and engaging.
Overall, the experiments demonstrate the advantages of the ensemble approach, where many different models output natural language responses and the system policy selects one response among them.
With more interactions and data, the learned policies are bound to continue improving.

\section{Conclusion}
We have proposed and evaluated a new large-scale ensemble-based dialogue system framework for the Amazon Alexa Prize competition.
Our system leverages a variety of machine learning methods, including deep learning and reinforcement learning.
We have developed a new set of deep learning models for natural language retrieval and generation, including deep learning models.
Further, we have developed a novel reinforcement learning procedure and evaluated it against existing reinforcement learning methods in A/B testing experiments with real-world Amazon Alexa users.
These innovations have enabled us to make substantial
improvements upon our baseline system.
Our best performing system reached an average user score of $3.15$, on a scale $1-5$, with a minimal amount of hand-crafted states and rules and without engaging in \textit{non-conversational activities} (such as playing games).
In comparison, the average user score for all teams in the competition during the semi-finals was only $2.92$.
Furthermore, the same system averaged $14.5-16.0$ turns per conversation, which is substantially higher than the average number of turns per conversation of all the teams in the semi-finals.
This improvement in back-and-forth exchanges between the user and system suggests that our system is one of the most \textit{interactive} and \textit{engaging} systems in the competition.
Since nearly all our system components are trainable machine learning models, the system is likely to improve greatly with more interactions and additional data.

\subsubsection*{Acknowledgments}

We thank Aaron Courville, Michael Noseworthy, Nicolas Angelard-Gontier, Ryan Lowe, Prasanna Parthasarathi and Peter Henderson for helpful feedback.
We thank Christian Droulers for building the graphical user interface for text-based chat.
We thank Amazon for providing Tesla K80 GPUs through the Amazon Web Services platform.
Some Titan X GPUs used for this research were donated by the NVIDIA Corporation.
The  authors  acknowledge  NSERC,  Canada  Research  Chairs,  CIFAR,  IBM  Research,  Nuance Foundation, Microsoft Maluuba and Druide Informatique Inc. for funding.



\small
\bibliographystyle{abbrvnat}
\bibliography{papers}



\end{document}